# Asymmetric Attributed Embedding via Convolutional Neural Network

Mohammadreza Radmanesh[1, *], Hossein Ghorbanzadeh[2, *], Ahmad Asgharian Rezaei[1], Mahdi Jalili[1 #], Xinghuo Yu[1]

*1 School of Engineering, RMIT University, Melbourne, Australia*
*2 Department of Computer Engineering, Ahvaz Branch, Islamic Azad University, Ahvaz, Iran*

**Abstract:** Recently network embedding has gained increasing attention due to its advantages in facilitating network computation tasks such as link prediction, node classification and node clustering. The objective of network embedding is to represent network nodes in a low-dimensional vector space while retaining as much information as possible from the original network including structural, relational, and semantic information. However, asymmetric nature of directed networks poses many challenges as how to best preserve edge directions in the embedding process. Here, we propose a novel deep asymmetric attributed network embedding model based on convolutional graph neural network, called AAGCN. The main idea is to maximally preserve the asymmetric proximity and asymmetric similarity of directed attributed networks. AAGCN introduces two neighbourhood feature aggregation schemes to separately aggregate the features of a node with the features of its in- and out- neighbours. Then, it learns two embedding vectors for each node, one source embedding vector and one target embedding vector. The final representations are the results of concatenating source and target embedding vectors. We test the performance of AAGCN on three real-world networks for network reconstruction, link prediction, node classification and visualization tasks. The experimental results show the superiority of AAGCN against state-of-the-art embedding methods.

## 1. Introduction

Real-world networked systems are growing in ubiquity by recent advances in the World Wide Web infrastructure and mobile communication devices [1]. The global tendency in making the world increasingly interconnected leads to emergence of large-scale networks such as citation networks, protein-protein interaction (PPI) networks, social networks, power grids, telecommunication networks and biological networks [2]. Adjacency matrix is a common approach to represent relationships in a network explicitly using a set of connected pairs of nodes [3]. However, the complex topological characteristics of these networks such as sparsity and directed edges along with meta information associated to each node, i.e., node features, bring challenges to network computational tasks such as link prediction, node classification and network reconstruction [4-6]. Recently more works in the literature have focused on learning informative node representations [7, 8], which is also referred to as network embedding methods.

Network embedding aims at projecting the original network into a low-dimensional vector space while maximally preserving meaningful structural, relational, and semantic information among network nodes. Network embedding methods are mainly grouped into three categories: 1) random walk-based methods such as node2vec [9] and DeepWalk [10]; 2) matrix factorization-based methods such as GraRep [11] and Locally

\# Corresponding author
\* Contributed equally
E-mail addresses: m4.radmanesh@gmail.com (M.Radmanesh), hossein.ghorbanzadeh@iauahvaz.ac.ir (H.Ghorbanzadeh), a.asghariyrezayi@gmail.com (A.A. Rezaei), mahdi.jalili@rmit.edu.au (M.Jalili), xinghuo.yu@rmit.edu.au (X.Yu)
This research is funded by Australian Research Council through Project No. DP170102303.

Linear Embedding (LLE) [12], and 3) neural network based methods which itself can be divided into two subcategories of graph autoencoders (such as TDNE [13], AGE [14], and deep contractive AE [2]) and convolutional graph neural networks (CGNNs) (such as GraphSAGE [15]). Among these methods, methods that are based on neural networks can be applied on different types of networks for various downstream tasks. This is because that due to the universal approximation theory [16] neural networks can potentially approximate any non-linear function. Moreover, the recent advances in computer hardware and cloud computing made neural network models scalable too [17].

In this regard, the convolutional graph neural network is the generalized form of a convolution function with two dimensions where the graph convolutional operation is applied on a network structure instead of a two dimensional image [17]. It aims to generate representations of nodes by aggregating the features of the nodes and their neighbours. In fact, CGNN methods fall into two categories of spectral-based and spatial-based models. Spectral-based CGNNs are common in graph signal processing and are used for defining graph convolutions based on spectral relations among nodes. These approaches assume that the network is undirected and learn representations of nodes by applying multiple spectral filters on the network, followed by a nonlinear activation function. The graph convolutional operation in these methods intends to denoise the graph signals. Spatial-based CGNNs benefit the idea of information propagation of recurrent neural networks (RNNs) when defining the graph convolutional operation and learn node representations using the local spatial relations between a node and its neighbours.

Many real-world networks are directed. In a such network, an edge presents not only a connection between a pair of nodes but also direction of flow of information through the network. Edge direction plays a key role in determining the structural properties of the network, i.e., asymmetric proximity and asymmetric similarity. For example, in a social network, when two individuals follow the same persons, they most likely share common interests even if there is no explicit connection between them, but the reverse is not always true. In a directed network, disregarding directionality by symmetrizing the connection between network nodes results in vital asymmetric information loss, since the quality of node representations greatly depends on how well we preserve asymmetric proximity and similarity during the embedding process.

Studying asymmetric network embedding faces four major limitations. The first challenge is the scalability of the models [18]. With the rapid growth of real-world networks in size, information mining approaches needs to fit networks with more complex structure and meta content. Among them, methods based on matrix factorization such as GraRep often fail on large-scale networks due to their massive time and memory complexities. Model Input Attachment (MIP) is another major shortcoming in the literature which greatly affects the generalizability of the models to different types of networks and downstream tasks [19]. MIP reflects the dependency of a model to a specific similarity measure such as Katz Index [20], Adamic-Adar [21] and Common Neighbours when creating the input network. As an example, the performance of HOPE [22] varies a lot by swapping the similarity measure used in the model. Another consideration is distribution of nodes. On directed networks, a significant portion of nodes have in-degree or out-degree of zero [23] that results in inaccessibility between pairs of nodes. Ignoring the distribution of nodes gives rise to barriers in a wide range of approaches that somehow employ random walk during the embedding process including (a) methods that are literally based on random walk such as APP [24], (b) methods that rely on random walk objectives such as GraphSAGE, and (c) methods that use random walk to build the input network such as HOPE. This is because for nodes with in-degree or out-degree of zero, the node representations cannot be effectively trained. Considering the network of Figure 1 as an example, although in-degree of nodes D and A are zero, they have exactly the same neighbourhood structure with four mutual nodes out of six which implies the high degree of similarity between these two nodes. Giving this network to APP returns a similarity value of zero between nodes D and A. Similarly, the proximity measures of HOPE return a proximity of zero between these nodes. Last but not least, most asymmetric embedding models merely utilize topology of networks and ignore a valuable source of information, i.e., node attributes. However, methods that couple topology of networks with additional information result in better representations compared with the methods that merely preserving structure proximities [25].

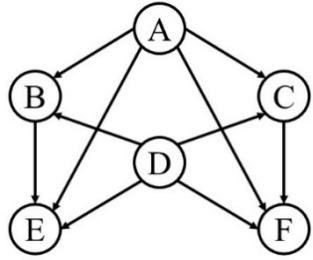

Figure 1. A toy example of a directed network.

In this work we propose a novel deep asymmetric attributed network embedding model, Asymmetric Attributed Graph Convolutional Network (AAGCN), that preserves the asymmetric proximity and asymmetric similarity of directed attributed networks during the embedding process. AAGCN, as the name implies, is based on CGNNs that means it possesses all properties of neural network-based embedding models in addressing the non-linear structure of a network and the model scalability. Figure 2 illustrates the framework of AAGCN. To preserve the Asymmetric characteristics of directed attributed networks, first, the proposed method separately aggregates the features of a node with the features of its in- and out- neighbours. The neighbourhood features aggregation schemes address the challenge of node distribution (inaccessibility between pairs of nodes) by enriching the node representations. Then, it learns two embedding vectors for each node, one source embedding vector and one target embedding vector. At the final step, the model provides an asymmetric representation for each node by concatenating the corresponding learned source and target embeddings.

The key contributions of this paper are as follows:

- To the best of our knowledge, this is the first CGNN-based model that could preserve asymmetric proximity and similarity on directed attributed networks. The optimal order of asymmetric similarity is selected based on networks and target tasks.
- The model only needs the adjacency matrix of the network and node attributes to learn the representations. This implies that AAGCN has no dependency on a strict similarity measure when creating the input network, indicating the generalizability of AAGCN to a wide range of networks and downstream tasks.
- The neighbourhood features aggregation schemes address the challenge of node distributions by enriching the node representations.
- Most of the CGNN-based embedding methods are structured in the form of a classifier [26] but our method goes further and verify the performance of learned representations on link prediction, network reconstruction and visualization tasks as well.

The organization of this paper is described below. In Section 2, we state the definition of the problem. Section 3 reviews the related works on directed network embedding. In section 4, the proposed AAGCN framework is detailed. Section 5 introduces the used datasets and discusses the results, followed by the conclusion and future directions in section 6.

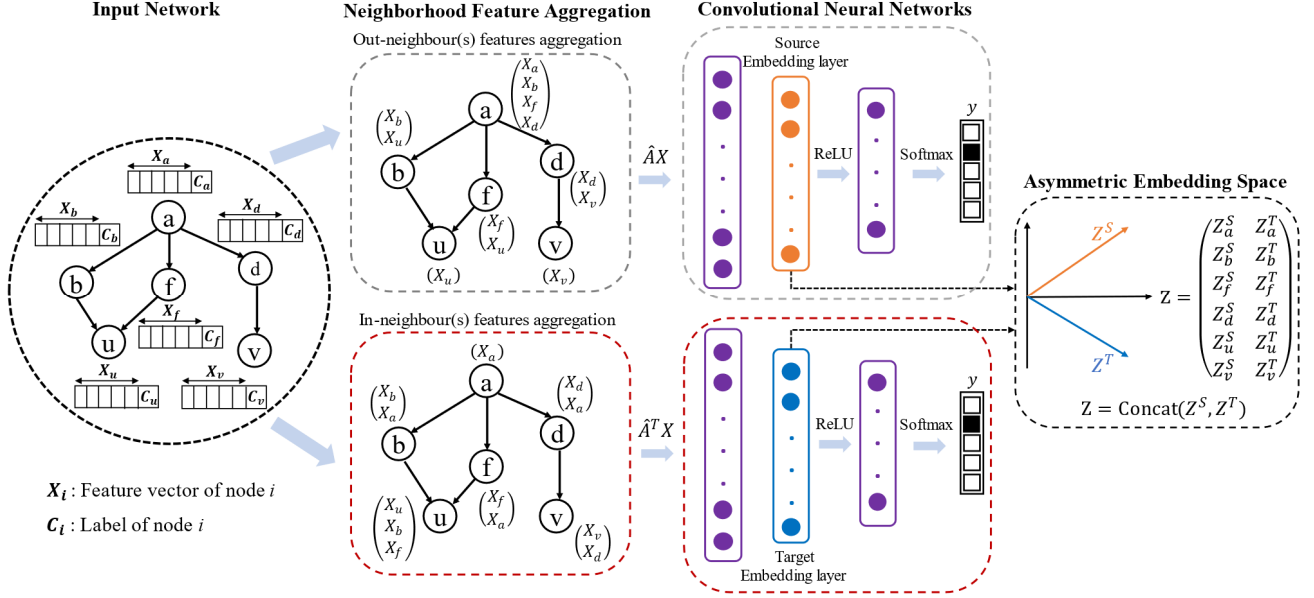

Figure 2. Framework of AAGCN model. For node $i$, $X_i$ and $C_i$ denote feature vector and class, respectively. The model, first, separately aggregates the features of a node with the features of its in- or out- neighbour(s). Then, for each node, it learns two separate low-dimensional representations (source embedding vector and target embedding vector).To obtain the final embedding of the node, it concatenates the corresponding source and target embeddings.

## 2. Problem definition

Let's define a directed attributed graph as $G(V, E, A, X, C)$ where $V = \{v_1, v_2, \ldots, v_n\}$ and $E = \{e_{i,j}\}_{i,j=1}^n$ specify the set of nodes with the size of $n$ and the set of edges respectively. $e_{i,j} = (v_i, v_j)$ represents a directed connection from node $v_i$ to node $v_j$. $A \in \mathbb{R}^{n \times n}$ denotes the adjacency matrix where $a_i = [a_{i,1}, a_{i,2}, \ldots, a_{i,n}]$ represents the $i$-th row of $A$ that includes non-negative weights of edges; $a_{i,j} = 1$ if there is a directed connection from $v_i$ to $v_j$, and $a_{i,j} = 0$ otherwise. $X \in \mathbb{R}^{n \times F}$ defines the attribute matrix where $F$ is the number of attributes associated to each node. $C$ specify the node label vector.

***Definition 1.*** (Source and target node): In a directed network, an edge presents not only the explicit connection between a pair of nodes but also direction of the connection. In this way, considering the edge $e_{i,j}$, node $v_i$ is the source node and node $v_j$ is the target node.

***Definition 2.*** (Asymmetric proximity): reflects the connection strength of a pair of nodes in a directed network. For nodes $v_i$ and $v_j$, $a_{i,j}$ represents the asymmetric proximity of $(v_i, v_j)$ and $a_{j,i}$ represents the asymmetric proximity of $(v_j, v_i)$ where $a_{i,j} \not\equiv a_{j,i}$.

***Definition 3.*** (Asymmetric similarity): reflects the neighbourhood similarity of a pair of nodes in a directed network. For nodes $v_i$ and $v_j$, $Asim_{i,j}$ and $Asim_{j,i}$ represent the asymmetric neighbourhood similarity of $(v_i, v_j)$ and $(v_j, v_i)$, respectively, where $Asim_{i,j} \not\equiv Asim_{j,i}$. This definition is not limited merely to the structural neighbourhood similarity of a pair of nodes, but it incorporates the node feature similarity as well.

***Definition 4.*** (Asymmetric network embedding): aims at representing the original high-dimensional network in a low-dimensional vector space so that asymmetric characteristics (asymmetric proximity and similarity) are preserved. Mathematically, given a directed attributed network $G(V, E, A, X, C)$, asymmetric network embedding learns the embedding vector space by a mapping function $F: A \in \mathbb{R}^{n \times n} \times X \in \mathbb{R}^{n \times F} \to Z \in \mathbb{R}^{n \times 2d}$, where $d \ll n$. The embeddings are represented as $Z = ((z_1^S. z_1^T), (z_2^S. z_2^T), \ldots, (z_n^S. z_n^T))$, where $z_i^S \in \mathbb{R}^d$ and $z_i^T \in \mathbb{R}^d$ define source embedding vector and target embedding vector of $v_i$ respectively. The embedding of $v_i$, $Z_i = (z_i^S. z_i^T) \in \mathbb{R}^{2d}$, is obtained by concatenating $z_i^S$ and $z_i^T$.

We summarize the terms and notations in Table 1.

Table 1. Terms and notations

| Symbols | Definition |
|---|---|
| $|V| = n$ | Number of network nodes |
| $F$ | Number of features associated to each node |
| $l$ | Number of hidden layers |
| $L$ | Number of layers |
| $|C|$ | Number of classes |
| $A \in \mathbb{R}^{n \times n}$ | Adjacency matrix of network |
| $X \in \mathbb{R}^{n \times F}$ | Attribute matrix of network |
| $d, \ d \ll n$ | Embedding size |
| $z_i^S$ | Source embedding vector of node $v_i$ |
| $z_i^T$ | Target embedding vector of node $v_i$ |
| $Z_i = (z_i^S . z_i^T) \in \mathbb{R}^{2d}$ | Embedding vector of node $v_i$ |

# 3. Related works

In the last decade, network embedding technology has significantly advanced due to its applications in graph-mining methods. Network embedding methods can be categorized into three groups including 1) random walk-based methods 2) matrix factorization-based methods and 3) neural network-based methods. Here we focus on related works that have been proposed for directed networks and/or incorporate node features.

## 3.1. Matrix Factorization

Factorization based methods derive input matrix based on node pairwise similarity and obtain the node representations by factorizing this matrix. HOPE [22] constructs the input similarity matrix of directed network by approximating four proximity measures, e.g., Common Neighbours, Rooted PageRank [27], Katz Index and Adamic-Adar and then decompose this matrix using a generalized Singular Value Decomposition (SVD). The main limitations of methods in this category are their model scalability and generalizability.

## 3.2. Random Walk

Random walks are shallow embedding approaches used to embed nodes into similar vectors if they tend to co-occur on short paths (i.e., random walks) over the graph. APP proposes an asymmetric embedding method via random walk with restart on directed networks that preserve Rooted PageRank proximity [28]. Attri2Vec [29] attempts to increase the consistency of network topology with node features by introducing a latent attribute subspace. NERD [19] is an asymmetric approach that samples neighbours of a node using an alternating random walk strategy. ANSE [30] and LRAP [31] are other asymmetric embedding models based on random walk.

## 3.3. Graph Neural Networks

The recent advances in graph neural networks (GNNs) have revolutionized many machine learning tasks. GNNs can be roughly divided into two categories: graph autoencoders and convolutional graph neural networks. Graph autoencoder includes an encoder that compresses the input network into a representation space and a decoder that reconstruct the input network back from embeddings. Gravity-inspired [32] introduces an asymmetric network embedding method inspired by newton's gravity theory. Gravity-inspired incorporates the topological information of networks for directed link prediction task. Convolutional graph neural network (GCNN) is a variant of convolutional neural networks [33] most commonly applied to computer vision. GCN [34] by making various simplifications and approximation addresses the computational complexity of spectral CGNNs and propose a semi-supervised classifier using topology of the network and node features. The unsupervised model of graphSAGE [35] obtains a node representation by aggregating features of a fixed number of its neighbours. DGI [36] focuses on capturing global topological information by maximizing local mutual information. RGCN

[37] propose an encoder framework for entity classification and link prediction on heterogenous networks. Table 2 provides more details on the related works.

Table 2. Summary of related works. L: the number of layers, |E|: the number of non-zeros in adjacency matrix, P: the number of neighbours per node, F: the number of features, b: represents the batch size, I: the iteration number, S: the singular value, R: the number of samples per node, N: the sample length, d: the embedding size, and K: the number of negative samples.

| Models | Category | Support Directed Networks | Asymmetric Embedding | Support Node Features | Preserve Asymmetric Properties | Time Complexity | Space Complexity | Limitations |
|---|---|---|---|---|---|---|---|---|
| HOPE [22] | Matrix Factorization | ✗ | ✗ | ✓ | ✗ | $O(\|E\|S^2 I)$ | $O(\|E\|S^2)$ | Scalability, MIP, Ignore node distribution, Disregard meta data |
| Attrib2vec [29] | Random Walk | ✗ | ✗ | ✓ | ✗ | $O(\|E\|d\|V\|)$ | $O(\|E\|d\|V\|)$ | Ignore node distribution |
| APP [28] | Random Walk | ✓ | ✓ | ✗ | ✓ | $O(\|V\|dRNI)$ | $O(2\|V\|d + \|E\|)$ | Ignore node distribution, Disregard meta data |
| NERD [19] | Random Walk | ✓ | ✓ | ✗ | ✓ | $O(Kdl\|V\| + \|E\|)$ | $O(\|E\|)$ | Ignore node distribution, Disregard meta data |
| GCN [34] | CGNN | ✗ | ✗ | ✓ | ✗ | $O(L\|E\|F + L\|V\|F^2)$ | $O(L\|V\|F + LF^2)$ | Sym-Embedding |
| graphSAGE [35] | CGNN | ✗ | ✗ | ✓ | ✗ | $O(P^L\|V\|F^2)$ | $O(bP^L F + LF)$ | Ignore node distribution |
| RGCN [37] | CGNN | ✓ | ✗ | ✓ | ✗ | - | - | Sym-Embedding |
| DGI [36] | CGNN | ✗ | ✗ | ✓ | ✗ | - | - | Sym-Embedding |

# 4. Proposed method

In this section, we introduce the proposed asymmetric spatial graph convolution (AAGCN). We start by introducing our model with single hidden layer and then expand it to the deep structure. We also discuss time and space complexities of the proposed model.

## 4.1. Asymmetric Spatial Graph Convolution Method

In a directed attributed network, we expect network nodes to be mapped close to each other based on not only their structural similarities but also their attribute similarities. Involving the node attributes during the embedding process, adds more informative information to the node representations that prevents the similar nodes with in- or out- degree of zero to be mapped faraway in the embedding space. Spatial CGNNs define a neighbourhood feature aggregation scheme for each node based on the pairwise local spatial relationships. An obvious solution for feature aggregation is to take the average value of the features of node $v_i$ along with its neighbours. However, in a directed graph, this approach disregards the direction of edges. In a directed graph, in- and out- neighbours of node $v_i$ are defined as follow:

$$\Gamma_{in}^i = \{j | (j, i) \in E\} \quad (1)$$

$$\Gamma_{out}^i = \{j | (i, j) \in E\} \quad (2)$$

where $|\Gamma_{in}^i|$ and $|\Gamma_{out}^i|$ are in-degree and out-degree of node $v_i$ respectively.

Here, to preserve the asymmetric nature of directed networks (i.e., asymmetric proximity and asymmetric similarity between nodes), we propose two distinct feature aggregation schemes used in source and target embedding processes. To this end, we simultaneously train two graph convolutions by passing them the adjacency matrix $A$ and the attribute matrix $X$ of the network. Considering the graph convolutions with only one hidden layer, in- and out- neighbour(s) feature aggregations of node $v_i$ are defined as follows:

$$h_i^S = \sigma\left(\sum_{j\epsilon\{\Gamma_{out}^i \cup i\}} X_j W^{S^{(0)}}\right) \quad (3)$$

$$h_i^T = \sigma\left(\sum_{j\epsilon\{\Gamma_{in}^i \cup i\}} X_j W^{T^{(0)}}\right) \quad (4)$$

where $W^{S^{(0)}}$ and $W^{T^{(0)}}$ are the learnable weight matrix for the single hidden layer of source and target graph convolutions, respectively, and $\sigma(.)$ is a non-linear activation function. $h_i^S \in \mathbb{R}^d$ represents the output of source graph convolution that aggregates the features of node $v_i$ along with its out-neighbour(s). Similarly, $h_i^T \in \mathbb{R}^d$ represents the output of target graph convolution that aggregates the features of node $v_i$ along with its in-neighbour(s). Involving the features of a node during the feature aggregation process, addresses the challenge of distribution of nodes on directed networks. In detail, when the number of in- or out- neighbour(s) of node $v_i$ are low or even zero, considering the features of the node $v_i$ itself during learning representations helps the pair of nodes including node $v_i$ with similar features to be mapped close to each other in the embedding space, even when there is no explicit edge between them in the input space. The $i$-th row of $A$ includes the out-neighbour(s) of node $v_i$ and its $i$-th column includes the in-neighbour(s) of node $v_i$. Defining $\hat{A} = A + I_n$ as adjacency matrix of network with added self-connections ($I_n$ is the identity matrix), Eqs. (3) and (4) can be factored as follows:

$$H^S = \sigma(\hat{A} X W^{S^{(0)}}) \quad (5)$$

$$H^T = \sigma(\hat{A}^T X W^{T^{(0)}}) \quad (6)$$

where $H^S \in \mathbb{R}^{n \times d}$ and $H^T \in \mathbb{R}^{n \times d}$ are source and target representations, respectively, and $\hat{A}^T$ is the transpose of $\hat{A}$. The final asymmetric representations are obtained by concatenating $H^S$ and $H^T$:

$$Z = concat(H^S, H^T) \quad (7)$$

We consider $ReLU(.) = \max(.,0)$ as activation function of the hidden layer. In addition, we structure each graph convolutions in the form of a classifier with $Softmax(h_i) = exp(h_i)/\sum_i exp(h_i)$ activation function in the output layer and set the number of neurons in this layer to the number of classes $|C|$. We update learnable weight matrices by back-propagation algorithm [38]. The outputs of the graph convolutions are defined as follows:

$$Y^S = softmax\left(\hat{A}\ ReLU\left(\hat{A} X W^{S^{(0)}}\right) W^{S^{(1)}}\right) \quad (8)$$

$$Y^T = softmax\left(\hat{A}^T ReLU\left(\hat{A}^T X W^{T^{(0)}}\right) W^{T^{(1)}}\right) \quad (9)$$

where $Y^S \in \mathbb{R}^{n \times |C|}$ and $Y^T \in \mathbb{R}^{n \times |C|}$ are node labels.

Increasing the number of hidden layers (i.e., convolutional layers) to $l$, extracts $l$-th order similarity among nodes. This is because of weight sharing over the whole network,. The general forms of feature aggregation schemes with $l$ layer are:

$$H^{S^{(l)}} = \sigma\left(\hat{A}\, H^{S^{(l-1)}} W^{S^{(l-1)}}\right) \quad (10)$$

$$H^{T^{(l)}} = \sigma\left(\hat{A}^T H^{T^{(l-1)}} W^{T^{(l-1)}}\right) \quad (11)$$

where $H^{S^{(0)}} = H^{T^{(0)}} = X$. If we represent the output of last hidden layer of source and target graph convolutions as $Z^S$ and $Z^T$, the overall asymmetric representations are represented as $Z = concat(Z^S, Z^T)$. Finally, to minimize the classification errors, we use cross-entropy loss function in graph convolutions:

$$\mathcal{L}_S = -\sum_{c \in C} \sum_{i=1}^{n} Y_i^c \ln(Y_i^S) \quad (12)$$

$$\mathcal{L}_T = -\sum_{c \in C} \sum_{i=1}^{n} Y_i^c \ln(Y_i^T) \quad (13)$$

where $Y_i^S$ and $Y_i^T$ are predicted labels of node $v_i$ in source and target graph convolutions respectively and $Y_i^c$ is the actual label of node $v_i$. $\mathcal{L}_S$ and $\mathcal{L}_T$ represents the classification errors of source and target graph convolutions respectively.

### 4.2. Memory and Time Complexities

We train two semi-supervised convolutional classifiers and concatenate the outputs of their last hidden layers to obtain the representations. Storing the adjacency matrix $A$ in the form of a sparse matrix results in the model overall memory complexity of $2L|E|F \in O(|E|F)$ where $L$ is the number of network layers and $|E|$ represents the number of non-zero elements of $A$. Most real-world networks are highly sparse which means only a small portion of $A$'s elements are non-zero. The linear memory complexity exposes the scalability of the proposed method. Similarly, the computational complexity of the model depends on the number of convolutional operations. We utilize the sparse matrix multiplication to calculate the convolutional operations. Therefore, we can obtain the computational complexity of AAGCN as $2L|E|F|K| \in O(|E|F|K|)$.

## 5. Results and discussion

We conduct experiments to assess the effectiveness of AAGCN against five state-of-the-art baselines embedding algorithms on four downstream tasks including link prediction, network reconstruction, node classification and network visualisation. Results show the superiority of our method in addressing the sparsity problem in link prediction and network reconstruction tasks and competitive advantages on node classification and network visualization tasks.

### 5.1. Datasets

We investigate the performance of representations on three real world academic citation networks. The nodes are scientific papers and directed connections are citations relations between them. All datasets are publicly available.

**Cora** [39] is composed of 2708 academic publications with 5429 citations relations in seven categories. Each publication is represented by a binary vector of 1433 words, in accordance with presence/absence of the corresponding word.

**DBLP** [40] consists of four publication scopes including Data Mining, Database, Artificial Intelligence and Computer Vision. It contains 18448 papers and 45661 citations. each paper is cited at least once and described by a binary vector of 2476 words.

**PubMed** [41] includes 19717 scientific papers in diabetes from three categories. There exist 44338 citations among publications. Each paper is represented by a TF-IDF weighted vector of 500 words.

Statistics of the networks are summarized in Table 3.

Table3 . Statistics of the real-world networks

| Network | Type | # Nodes | # Edges | # Feature | # Classes |
|---------|------|---------|---------|-----------|-----------|
| Cora | Directed | 2708 | 5429 | 1433 | 7 |
| Citeseer | Directed | 3312 | 7432 | 3703 | 6 |
| PubMed | Directed | 17717 | 44338 | 500 | 3 |
| DBLP | Directed | 18448 | 45661 | 2476 | 4 |

In directed networks, in- or out- degree of nodes highly impacts the performance of network embedding algorithms since there is no path between nodes with in- or out- degree of zero. In methods based on random walk, this fact results in a similarity value of zero for a significant portion of node pairs. Figure 3 shows in/out degree distribution of networks on scale of log-log. DBLP and Cora networks have the highest and the lowest number of nodes with in- or out- degree of zero, respectively.

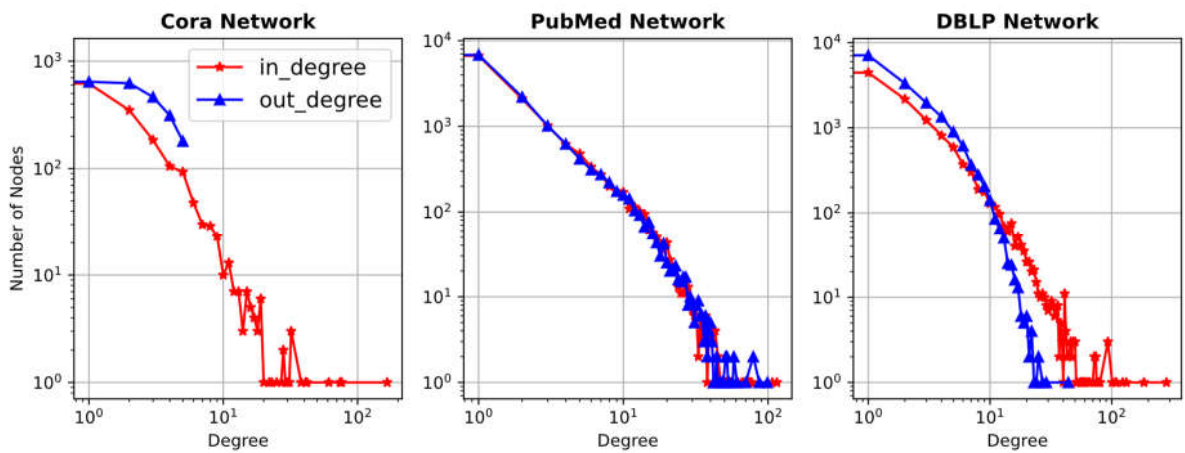

Figure 3. In/out degree distribution of networks on scale of log-log.

## 5.2. Experimental Setup

The proposed method is developed in Python software using TensorFlow, Numpy, StellarGraph, NetworkX and DGL packages. We compare the performance of our model with five baselines. Three methods are based on CGNN including GraphSAGE, DGI and RGCN. One method is based on random walk called Attri2Vec, and the other method is based on matrix factorization called HOPE. We train all two-layer CGNN-based methods on 100 epochs. In GraphSAGE we first calculate a normalized adjacency matrix $\hat{A} = \widetilde{D}^{-\frac{1}{2}} \widetilde{A} \widetilde{D}^{-\frac{1}{2}}$ where $\widetilde{A} = A + I$ represents the adjacency matrix with added self-connections and $\widetilde{D}_{ii} = \sum_j \widetilde{A}_{ij}$ is a diagonal matrix of node degrees. For baselines we set the hyperparameters suggested by the authors.

## 5.3. Asymmetric Similarity Measure

In a directed network, it is vital to preserve not only the connection between nodes but also their directions. Our method learns two different embeddings for each node, one based on the in-neighbours of the node (called target embedding vector), and another based on the out-neighbours (called source embedding vector). Here, we propose an asymmetric similarity measure of $(v_i, v_j)$ based on source and target embeddings to involve the link directions when measuring the similarity between pairs of nodes as follows:

$$S_{i,j} = dot(Z_i^S, Z_j^T) = \sum_{m=1}^{d} (Z_i^S)_m \times (Z_j^T)_m \qquad (14)$$

where $dot(Z_i^S, Z_j^T)$ represents the inner product of source embedding vector of node $v_i$ with target embedding vector of node $v_j$. According to this measure, the similarity of $(v_i, v_j)$ and $(v_j, v_i)$ may differ from each other (e.g., $S_{i,j} \not\equiv S_{j,i}$). We use this measure to rank pair of node representations when predicting new links or reconstructing the network.

### 5.4. Evaluation Metrics

To evaluate the performance of the embedding algorithms, we adapt *Precision@k* as evaluation metric for link prediction and network reconstruction tasks and use Micro-F1, Macro-F1 and Accuracy as evaluation metrics for node classification task.

***Precision@k*** measures the proportion of relevant retrieved edges in the top-k set. It is defined as follows:

$$Pr@k = \frac{1}{k} \sum_{m=1}^{k} \delta_m \qquad (15)$$

where in link prediction task $\delta_m = 1$ is $m$-th predicted pair of nodes in the hidden edges and in reconstruction task represents the $m$-th predicted pair of nodes in the observed pair of nodes. $\delta_m = 0$ otherwise.

***Macro-F1*** is the mean of label-wise F1 score for the classification task.

$$Macro - F1 = \frac{1}{C} \sum_{m=0}^{C} F1(m) \qquad (16)$$

where $j$ represents $j$-th label index and $K$ is number of classes.

***Micro-F1*** is the harmonic mean of Micro-precision and Micro-recall. Micro-precision and Micro-recall are calculated from the sum of true positives (TP), false positives (FP), and false negatives (FN) over all labels as follows:

$$Micro - F1 = \frac{2 * MicPr * MicRec}{MicPr + MicRec} \qquad (17)$$

where Micro-precision is defined as $MicPr = \frac{\sum_{m=0}^{C} TP(m)}{\sum_{m=0}^{C}(TP(m)+FP(m))}$ and $MicRec = \frac{\sum_{m=0}^{C} TP(m)}{\sum_{m=0}^{C}(TP(m)+FN(m))}$ represents Micro-recall.

**Accuracy** measures the proportion of true prediction to all predictions. It is defined as follows:

$$Accuracy = \frac{TP + TN}{TP + TN + FP + FN} \qquad (18)$$

where TP = true positives, TN = true negatives, FP = false positives, and FN = false negatives.

## 5.5. Results

In this section, first, we evaluate the quality of node representations learned by different network embedding methods in the task of network reconstruction to show their capability in preserving asymmetric structure of the networks. Next, we conduct link prediction and node classification experiments to compare the generalizability of network embedding methods.

**Network reconstruction** is a typical strategy to assess the quality of node representations, in terms of preserving informative information. It is to reconstruct the network from representations. When we deal with directed networks, the network connections are not the only information to be preserved during the embedding process, but we need to capture their direction. For this reason, our method separately learns in- and out- neighbourhood structures per each node. Then, we apply the asymmetric similarity measure (refer to Eq. (14)) on every pair of nodes to reconstruct the network using edges with the highest ranks. We use *Precision@k* on top-*k* pairs of nodes as evaluation metric for all methods. To test the performance of different methods, we report the average results of 10 independent runs.

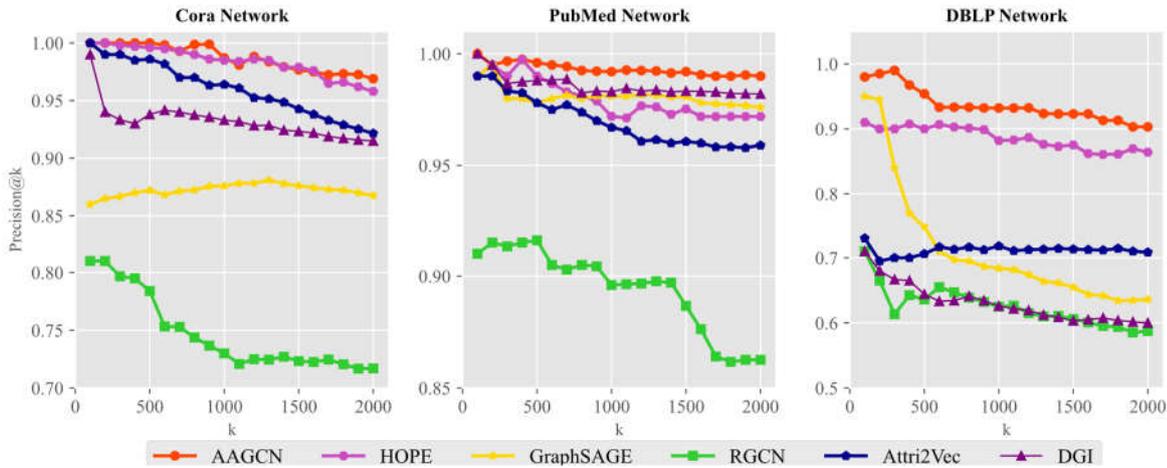

Figure 4. Network reconstruction results. Based on observations, AAGCN outperforms the baselines on PubMed and DBLP networks and except for HOPE that presents competitive performance outperforms other baselines on Cora network. We use a 3-layer network (two convolutional layers) for network reconstruction task.

Figure 4 shows *Pr@k* for different values of *k*. It shows that the proposed method, AAGCN, achieves superior results on PubMed and DBLP networks over the baselines in term of *Pr@k* metric. Similarly, on Cora network, except for HOPE that shows competitive results, AAGCN outperforms all other baselines. Still, AAGCN performs better than Hope on average, especially for the value of *k* higher than 1500. The observations demonstrate that the performances of random walk based methods highly depends on the distribution of nodes. For example, on DBLP network, which includes the highest percentage of nodes with in- or out- degree of zero or low, the performance of Attri2Vec is at least 25% worse than the other methods.. Similarly, the performance of GraphSAGE that relies on a loss function based on random walk, experiences a sharp drop by 32% as the number of pairs increases. In DBLP network, AAGCN almost maintains the performance over different number of pairs. This suggests the feature aggregation scheme of AAGCN could tackle the challenge of node distribution on directed networks by considering the feature similarities along with the structure proximities.

**Link Prediction** is a common way to investigate the generalizability of network embedding algorithms. Link prediction is used to examine the possibility of unobserved/missing links to occur in the future [42]. For a fair comparison, each result is avaraged over 10 independet runs. In each run, we randomly set aside 30% of links for testing and use the remaining 70% of links to train the model. Then, we rank all pairs of nodes useing the proposed asymmetric similarity measure (Eq. (14)). The top-ranking pairs, that are not in the training set, are selected as the future links. Same as network reconstruction task, we evelute the top-*k* pairs in terms of *Pr@k* metric.

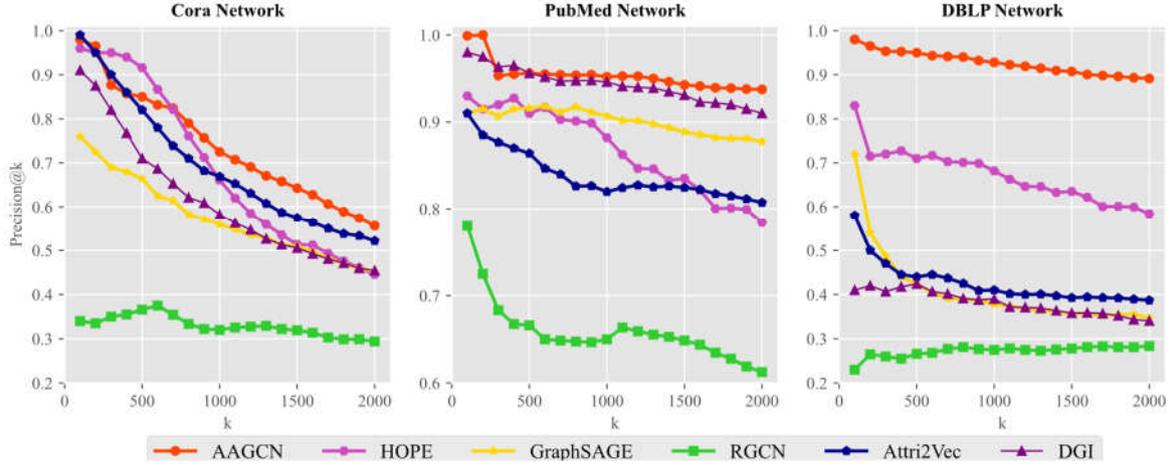

Figure 5. Link prediction results. Based on observations, AAGCN constantly outperforms the baselines on DBLP networks and by increasing *k* surpasses the baselines on Cora and PubMed networks. We use a 3-layer network (two convolutional layers) for link prediction task.

Figure 5 shows our proposed method outperforms the state-of-the-art algorithms by a margin of at least 5% on Cora network, and 3% on PubMed network, on average. The outperformance of AAGCN is much pronounced on DBLP network where it shows at least 26% better performance than other algorithms.. On Cora network, by reaching the number of pairs to 2000 the performance significantly drops since almost no other positive edge is remained to be predicted. Unlike network reconstruction task that the performance of HOPE was almost the same as AAGCN, in the link prediction task and for the values of *k* less than 700, HOPE presents better results that AAGCN but when *k* increases AAGCN significantly outperforms HOPE. In PubMed network, when $k < 1000$, DGI provides competitive results with AAGCN but by increasing *k*, AAGCN becomes the best performer. Similar to the network reconstruction task, ransom walk based methods gerenally show poor performance in the link prediction task. AAGCN, consistently and remarkably outperforms the baselines when the number of nodes with in- or out-degree of zero is high. Across all networks, RGCN provides the lowest effectiveness.

**Node Classification** task is used as another task to test the performance of the embedding algorithms. or a protion of nodes, the class to which they blonge may be unkwon; such nodes are refferred to as unlabelled nodes. Node classification refers to the task of assigning unlabelled nodes to a class with respect to the thier similarity to the nodes with known class (called labelled nodes). In this work, first, we randomly sample 80% of the learned representations to train a Logistic Regression [43] as a classifier and predict the class of remaining 20% node representations. The reported results are average of 10 times experiments.

Table 4. Results of node classification task in terms of Micro-F1, Macro-F1 and Accuracy. Based on the results, AAGCN outperforms the baseline algorithms. We use a 2-layer network (single convolutional layer) for node classification task. The best-performer is denoted by bold face.

|  |  | AAGCN | HOPE | GraphSAGE | RGCN | Atrri2Vec | DGI |
|---|---|---|---|---|---|---|---|
| Cora | Micro-F1 | **0.81 (0.02)** | 0.52 (0.05) | 0.63 (0.08) | 0.25 (0.04) | <u>0.80 (0.03)</u> | <u>0.80 (0.02)</u> |
|  | Macro-F1 | **0.80 (0.05)** | 0.46 (0.01) | 0.73 (0.04) | 0.22 (0.04) | <u>0.78 (0.08)</u> | 0.75 (0.07) |
|  | Accuracy | **0.83 (0.01)** | 0.52 (0.03) | 0.73 (0.04) | 0.22 (0.04) | <u>0.82 (0.08)</u> | 0.78 (0.1) |
| PubMed | Micro-F1 | **0.84 (0.01)** | 0.66 (0.04) | 0.79 (0.08) | 0.61 (0.05) | <u>0.83 (0.09)</u> | 0.82 (0.03) |
|  | Macro-F1 | **0.83 (0.02)** | 0.55 (0.07) | 0.79 (0.03) | 0.64 (0.07) | 0.81 (0.07) | <u>0.82 (0.05)</u> |
|  | Accuracy | **0.85 (0.02)** | 0.66 (0.07) | 0.78 (0.08) | 0.63 (0.08) | <u>0.83 (0.03)</u> | 0.82 (0.04) |
| DBLP | Micro-F1 | **0.83 (0.01)** | 0.61 (0.02) | <u>0.80 (0.02)</u> | 0.48 (0.05) | 0.80 (0.03) | **0.83 (0.03)** |
|  | Macro-F1 | **0.79 (0.02)** | 0.49 (0.01) | 0.73 (0.06) | 0.24 (0.03) | 0.74 (0.02) | <u>0.78 (0.01)</u> |
|  | Accuracy | **0.83 (0.08)** | 0.60 (0.05) | <u>0.80 (0.03)</u> | 0.45 (0.03) | 0.80 (0.07) | **0.83 (0.05)** |

Table 4 summarizes the comparison between our proposed method and baselines for classification task in terms of *Micro-F1*, *Macro-F1* and *Accuracy*. The proposed method outperforms the baseline algorithms accross all networks. DGI and Attri2Vec, which consider features of nodes during the embedding but ignore the directionality, presents reasonably competetive performance with AAGCN (The former is based on CGNNs and the latter is a method based on random walk). The results suggest that considering the node features may have more impact than the direction of edges on node classification task.

**Visualization** is the last task used to compare the algorithms. Projecting high-dimensional sparse networks into a 2-dimansional space is not informative. Network embedding enables us to map such networks into a meaningful low-dimensional space that can be used for network visualization purposes. To evaluate the visualization ability of AAGCN and compare it with the baselines, we map them into a 2-dimensional space by feeding their representations as input to *t-SNE* [44] on PubMed network. Samples with the same colour come from the same class. The effectiveness of visualization task depends on two aspects: *i*) group separation that is how far away classes are from each other and, and *ii*) group boundary that is how close samples are to each other within a class. To measure group separation and boundary we use Silhouette [45] score which is a value between -1 to +1 where high value indicates that samples are well assigned to their own groups and poorly assigned to other groups.

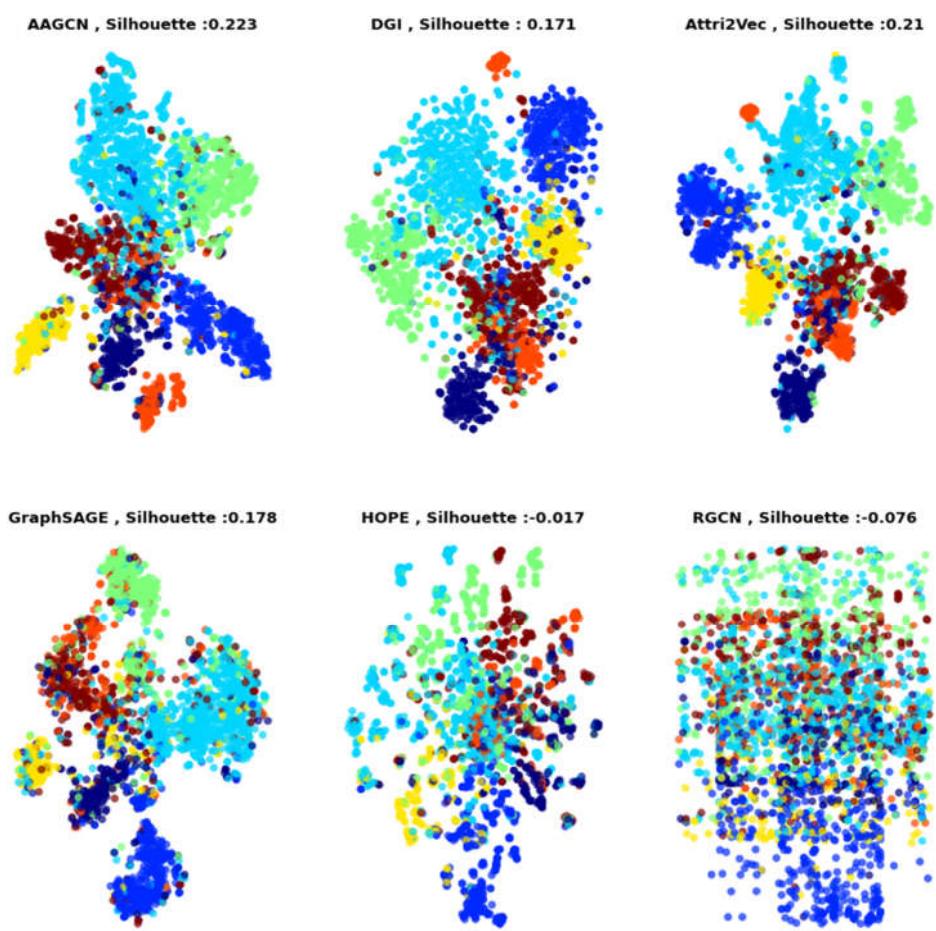

Figure 6. Visualization of node representations on PubMed network. Based on the visualization, AAGCN obtain the highest Silhouette among all methods.

Figure 6 shows that AAGCN achieves the best clustering performance among all methods with the Silhouette score of 0.223, followed by Attri2Vec with Silhouette score of 0.210. The layout of RGCN and HOPE are not satisfactory since samples of different colours are mixed with each other. For DGI and GraphSAGE, although samples of the same colour tend to form separate groups, the boundary between them is not clear.

## 5.6. Model Depth Analysis

In this section, we present the influence of different orders of similarity on the performance of network reconstruction, link prediction and node classification tasks by varying the number of hidden layers (i.e., convolutional layers) from 1 to 10. For simplicity, we only report the results of model depth analysis on PubMed network. In each experiment, we train the model using Adam [46] optimizer with a learning rate of 0.01 for 200 epochs. For all networks we set the size of embedding to 100 (e.g., $d = 100$). On the link prediction task, we evaluate the effectiveness of the proposed method of different depth in terms of *Pr@200* when we randomly remove 30% of edges as test data. On node classification task, we report the performance of Logistic Regression in terms of *Accuracy*.

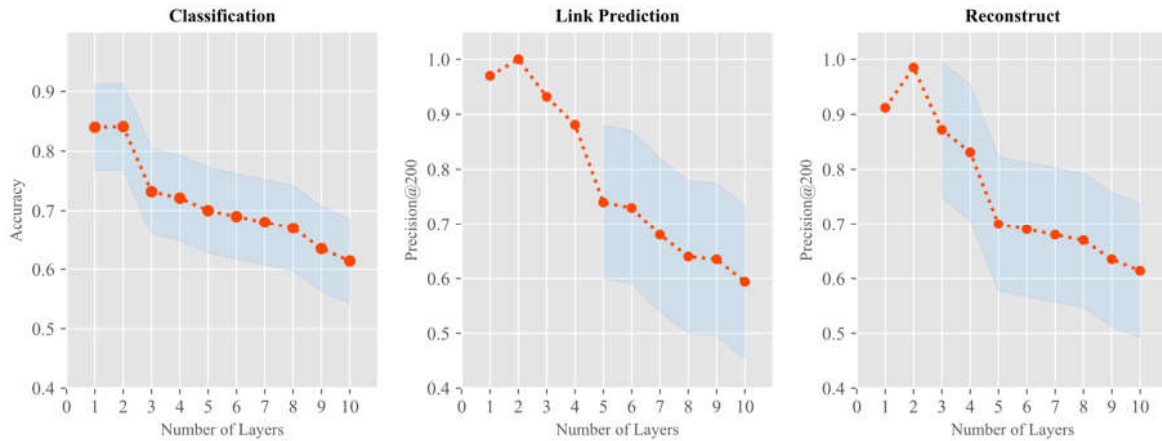

Figure 7. Results of model depth analysis.

When the model only has one hidden layer, it merely incorporates the 1st-order similarity between nodes. When the model has $l$ hidden layers, it considers $l$-th order similarities. However, the optimal similarity order greatly depends on networks and intended tasks which is often hard to be predetermined. Figure 6 presents the performance of different tasks when AAGCN goes deeper on PubMed network. Our model obtains the best results with one or two convolutional layers. For the number of convolutional layers more than two, the deeper the model, the lower the performance. This indicates that by increasing the number of hidden layers by more than two, the model starts to overfit to the training data. This is mainly due to the growing complexity that we impose to the model by adding every new layer.

## 6. Conclusion

In this paper, we introduced a novel asymmetric attributed network embedding model based on convolutional graph neural networks, called AAGCN, and validated it on three real-world networks including Cora, PubMed and DBLP networks. We demonstrated four underlying properties for AAGCN model. 1) It preserves asymmetric proximity and asymmetric neighbourhood similarity on directed attributed networks. The experiments showed that the optimal order of neighbourhood similarity highly depends on networks and target tasks. 2) It is not restricted to a specific similarity measure that highlights the generalizability of the model to different types of networks and downstream tasks. 3) It proposes two neighbourhood features aggregation schemes with respect to in- and out- neighbours of nodes that addresses the challenge of node distributions. On directed networks, a significant portion of nodes have no incoming or outgoing edges, indicating the inaccessibility between pairs of nodes. AAGCN introduces a neighbourhood similarity based on neighbourhood proximity and neighbourhood features. 4) AAGCN outperforms state-of-the-art baselines including GraphSAGE, RGCN, Attri2Vec, HOPE and DGI. The experimental results on downstream tasks including node classification, network reconstruction, visualization and link prediction showed the superior performance compared with baseline algorithms. Here we incorporate the network topology and node features to represent non-linear relations between network nodes. Future work would include generalizing the model to heterogenous and multilayer networks where heterogenous nodes can be connected through heterogenous links.

# References

1. Jalili, M., *Social power and opinion formation in complex networks.* Physica A: Statistical mechanics and its applications, 2013. **392**(4): p. 959-966.
2. Radmanesh, M., et al., *Online spike sorting via deep contractive autoencoder.* bioRxiv, 2021.
3. Asgharian Rezaei, A., M. Jalili, and H. Khayyam, *Influential Node Ranking in Complex Networks Using A Randomized DynamicsSensitive Approach.* arXiv e-prints, 2021: p. arXiv: 2112.02927.
4. Chen, J., et al., *Self-training enhanced: Network embedding and overlapping community detection with adversarial learning.* IEEE Transactions on Neural Networks and Learning Systems, 2021.
5. Lee, O.-J., H.-J. Jeon, and J.J. Jung, *Learning multi-resolution representations of research patterns in bibliographic networks.* Journal of Informetrics, 2021. **15**(1): p. 101126.
6. Li, Y., J. Yin, and L. Chen, *Seal: Semisupervised adversarial active learning on attributed graphs.* IEEE Transactions on Neural Networks and Learning Systems, 2020. **32**(7): p. 3136-3147.
7. Babaei, M., H. Ghassemieh, and M. Jalili, *Cascading failure tolerance of modular small-world networks.* IEEE Transactions on Circuits and Systems II: Express Briefs, 2011. **58**(8): p. 527-531.
8. Leskovec, J., J. Kleinberg, and C. Faloutsos, *Graph evolution: Densification and shrinking diameters.* ACM transactions on Knowledge Discovery from Data (TKDD), 2007. **1**(1): p. 2-es.
9. Grover, A. and J. Leskovec. *node2vec: Scalable feature learning for networks*. in *Proceedings of the 22nd ACM SIGKDD international conference on Knowledge discovery and data mining*. 2016.
10. Perozzi, B., R. Al-Rfou, and S. Skiena. *Deepwalk: Online learning of social representations*. in *Proceedings of the 20th ACM SIGKDD international conference on Knowledge discovery and data mining*. 2014.
11. Cao, S., W. Lu, and Q. Xu. *Grarep: Learning graph representations with global structural information*. in *Proceedings of the 24th ACM international on conference on information and knowledge management*. 2015.
12. Roweis, S.T. and L.K. Saul, *Nonlinear dimensionality reduction by locally linear embedding.* science, 2000. **290**(5500): p. 2323-2326.
13. Radmanesh, M., et al. *Topological Deep Network Embedding*. in *2020 International Conference on Artificial Intelligence in Information and Communication (ICAIIC)*. 2020. IEEE.
14. Khajehnejad, M., et al., *Adversarial graph embeddings for fair influence maximization over social networks.* arXiv preprint arXiv:2005.04074, 2020.
15. Hamilton, W., Z. Ying, and J. Leskovec, *Inductive representation learning on large graphs.* Advances in neural information processing systems, 2017. **30**.
16. Lu, L., et al., *Learning nonlinear operators via DeepONet based on the universal approximation theorem of operators.* Nature Machine Intelligence, 2021. **3**(3): p. 218-229.
17. Wu, Z., et al., *A comprehensive survey on graph neural networks.* IEEE transactions on neural networks and learning systems, 2020. **32**(1): p. 4-24.
18. Chang, S., et al. *Heterogeneous network embedding via deep architectures*. in *Proceedings of the 21th ACM SIGKDD international conference on knowledge discovery and data mining*. 2015.
19. Khosla, M., et al. *Node representation learning for directed graphs*. in *Joint European Conference on Machine Learning and Knowledge Discovery in Databases*. 2019. Springer.
20. Katz, L., *A new status index derived from sociometric analysis.* Psychometrika, 1953. **18**(1): p. 39-43.
21. Adamic, L.A. and E. Adar, *Friends and neighbors on the web.* Social networks, 2003. **25**(3): p. 211-230.
22. Ou, M., et al. *Asymmetric transitivity preserving graph embedding*. in *Proceedings of the 22nd ACM SIGKDD international conference on Knowledge discovery and data mining*. 2016.
23. Zhu, S., et al., *Adversarial directed graph embedding.* arXiv preprint arXiv:2008.03667, 2020.

24. Zhou, C., et al. *Scalable graph embedding for asymmetric proximity*. in *Proceedings of the AAAI Conference on Artificial Intelligence*. 2017.
25. Hou, M., et al., *Network embedding: Taxonomies, frameworks and applications.* Computer Science Review, 2020. **38**: p. 100296.
26. Chiang, W.-L., et al. *Cluster-gcn: An efficient algorithm for training deep and large graph convolutional networks*. in *Proceedings of the 25th ACM SIGKDD International Conference on Knowledge Discovery & Data Mining*. 2019.
27. Song, H.H., et al. *Scalable proximity estimation and link prediction in online social networks*. in *Proceedings of the 9th ACM SIGCOMM conference on Internet measurement*. 2009.
28. Shen, X. and F.-L. Chung. *Deep network embedding with aggregated proximity preserving*. in *Proceedings of the 2017 IEEE/ACM International Conference on Advances in Social Networks Analysis and Mining 2017*. 2017.
29. Zhang, D., et al., *Attributed network embedding via subspace discovery.* Data Mining and Knowledge Discovery, 2019. **33**(6): p. 1953-1980.
30. Dernbach, S. and D. Towsley, *Asymmetric Node Similarity Embedding for Directed Graphs*, in *Complex Networks XI*. 2020, Springer. p. 83-91.
31. Abu-El-Haija, S., B. Perozzi, and R. Al-Rfou. *Learning edge representations via low-rank asymmetric projections*. in *Proceedings of the 2017 ACM on Conference on Information and Knowledge Management*. 2017.
32. Salha, G., et al. *Gravity-inspired graph autoencoders for directed link prediction*. in *Proceedings of the 28th ACM international conference on information and knowledge management*. 2019.
33. Krizhevsky, A., I. Sutskever, and G.E. Hinton, *Imagenet classification with deep convolutional neural networks.* Advances in neural information processing systems, 2012. **25**.
34. Kipf, T.N. and M. Welling, *Semi-supervised classification with graph convolutional networks.* arXiv preprint arXiv:1609.02907, 2016.
35. Oh, J., K. Cho, and J. Bruna, *Advancing graphsage with a data-driven node sampling.* arXiv preprint arXiv:1904.12935, 2019.
36. Velickovic, P., et al., *Deep Graph Infomax.* ICLR (Poster), 2019. **2**(3): p. 4.
37. Schlichtkrull, M., et al. *Modeling relational data with graph convolutional networks*. in *European semantic web conference*. 2018. Springer.
38. Rumelhart, D.E., G.E. Hinton, and R.J. Williams, *Learning representations by back-propagating errors.* nature, 1986. **323**(6088): p. 533-536.
39. McCallum, A.K., et al., *Automating the construction of internet portals with machine learning.* Information Retrieval, 2000. **3**(2): p. 127-163.
40. Tang, J., et al. *Arnetminer: extraction and mining of academic social networks*. in *Proceedings of the 14th ACM SIGKDD international conference on Knowledge discovery and data mining*. 2008.
41. Namata, G., et al. *Query-driven active surveying for collective classification*. in *10th International Workshop on Mining and Learning with Graphs*. 2012.
42. Lü, L. and T. Zhou, *Link prediction in complex networks: A survey.* Physica A: statistical mechanics and its applications, 2011. **390**(6): p. 1150-1170.
43. Trouillon, T., et al. *Complex embeddings for simple link prediction*. in *International conference on machine learning*. 2016. PMLR.
44. Van der Maaten, L. and G. Hinton, *Visualizing data using t-SNE.* Journal of machine learning research, 2008. **9**(11).
45. Rousseeuw, P.J., *Silhouettes: a graphical aid to the interpretation and validation of cluster analysis.* Journal of computational and applied mathematics, 1987. **20**: p. 53-65.
46. Zhang, X., et al., *Identifying missing and spurious interactions in directed networks.* International Journal of Distributed Sensor Networks, 2015. **11**(9): p. 507386.